\documentclass[letterpaper, 10 pt, conference]{ieeeconf}  % Comment this line out if you need a4paper

\IEEEoverridecommandlockouts                              % This command is only needed if 
\usepackage{subcaption} % 导言区加上这个
\usepackage{algorithm}      % 提供 algorithm 环境和 caption
\usepackage{algorithmicx}   % 提供新的 algorithmic 环境
\usepackage{algpseudocode}  % algorithmicx 的伪代码风格，提供 \State, \Function 等
\usepackage{amsmath}        % 数学公式
\usepackage{amssymb}        % 数学符号
\usepackage{graphicx}
\usepackage{url} 
\newcommand{\mdp}{\mathcal{M}}
\newcommand{\sspace}{\mathcal{S}}
\newcommand{\aspace}{\mathcal{A}}
\newcommand{\saspace}{\mathcal{Z}}

\newcommand{\rawrew}{\mathcal{R}}

\newcommand{\rawtrns}{P}
\newcommand{\rew}{\rawrew}

\newcommand{\trns}{\rawtrns}

\newcommand{\dist}{{Dist}}

\title{Teaching RL Agents to Act Better: VLM as Action Advisor for Online Reinforcement Learning}

\author{Xiefeng Wu$^{1}$, Jing Zhao$^{1}$, Shu Zhang$^{1}$, Mingyu Hu$^{1}$%
\thanks{$^1$ Wuhan University, China. {\tt\small wuxiefeng@whu.edu.cn, mingyuhu@whu.edu.cn}}%
}

\begin{document}

\maketitle
\thispagestyle{empty}
\pagestyle{empty}

%%%%%%%%%%%%%%%%%%%%%%%%%%%%%%%%%%%%%%%%%%%%%%%%%%%%%%%%%%%%%%%%%%%%%%%%%%%%%%%%
\begin{abstract}
Online reinforcement learning in complex tasks is time-consuming, as massive interaction steps are needed to learn the optimal Q-function.Vision-language action (VLA) policies represent a promising direction for solving diverse tasks; however, their performance on low-level control remains limited, and effective deployment often requires task-specific expert demonstrations for fine-tuning.
In this paper, we propose \textbf{VARL} (\textbf{V}LM as \textbf{A}ction advisor for online \textbf{R}einforcement \textbf{L}earning), a framework that leverages the domain knowledge of vision-language models (VLMs) to provide action suggestions for reinforcement learning agents. Unlike previous methods, VARL provides action suggestions rather than designing heuristic rewards, thereby guaranteeing unchanged optimality and convergence. The suggested actions increase sample diversity and ultimately improve sample efficiency, especially in sparse-reward tasks.  
To validate the effectiveness of VARL, we evaluate it across diverse environments and agent settings. Results show that VARL greatly improves sample efficiency without introducing significant computational overhead. These advantages make VARL a general framework for online reinforcement learning and make it feasible to directly apply reinforcement learning from scratch in real-world environments.
\end{abstract}

\section{Introduction}

        % \vspace{-10pt}
\begin{figure*}[htbp]
    \centering
    \includegraphics[width=1\textwidth]{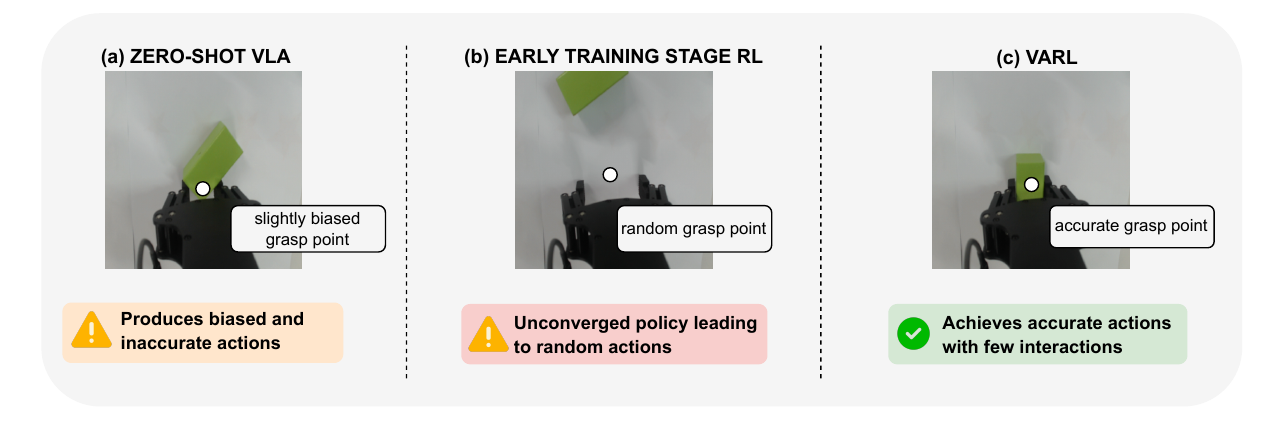}
        % \vspace{-10pt}
    \caption{\textbf{Illustrative comparison of different agents’ behaviors in a pickup task.} Zero-shot VLA tends to produce biased but inaccurate actions that only partially approach task success. Early-stage RL agents, before convergence, often exhibit near-random behaviors. In contrast, VARL, which integrates VLM-based heuristic actions with RL’s exploration, achieves accurate grasping with only a few interactions and converges efficiently.}
    \label{fig:core_varl}

\end{figure*}
% introduction section

Online reinforcement learning (RL) has shown great potential for achieving high performance without relying on expert demonstrations. However, one critical limitation of this approach is its \textit{sample inefficiency}\cite{alpha_go, google_agile_soccer}, as RL agents require extensive exploration to discover effective policies. In addition, the training process is often unstable.

To improve performance of agents on new tasks, two major directions have been explored. The first approach fine-tunes pretrained vision-language models (VLMs) using a few expert demonstrations \cite{interleav_vla,uniVLA,vla_lapa}. 
The second line of research leverages VLMs to construct reward functions for new tasks, followed by RL training \cite{rl-vlm-f, erl-rl}. In this setting, the VLM is queried to score pairs of trajectories for preference learning, and the resulting preferences are transformed into reward signals that guide policy optimization. Although this approach can maintain strong performance after convergence and eliminates the need for expert demonstrations, its effectiveness is constrained by the accuracy of the VLM.

Fine-tuning approaches can improve task success rates but still require the collection of expert demonstrations, while VLM-based reward shaping methods remove the need for expert data but rely on the assumption that VLM preferences are accurate. To overcome these limitations, we propose VARL, a novel online reinforcement learning framework that efficiently exploits the domain knowledge of VLMs. In VARL, VLMs serve as action advisors during training: given a state, the VLM suggests an action, and these suggestions are integrated into policy learning on a subset of sampled trajectories. This mechanism enriches exploration and enhances sample efficiency, while avoiding the drawbacks of both fine-tuning and reward-shaping methods.

Our experiments highlight three key advantages of VARL:

(1) \textbf{Generalization across diverse task settings.} VARL applies to both state-based and vision-based tasks, and supports discrete and continuous action spaces. By combining the high-level decision-making generalization of VLMs with the exploration ability of RL, it effectively leverages biased, suboptimal VLM information for efficient exploration.

(2) \textbf{Improved sample efficiency.} VARL achieves high performance with relatively few interaction steps, showing that online RL training under VLM supervision is a promising path toward reducing exploration cost.

(3) \textbf{Lower computational overhead.} Reward-shaping methods \cite{rl-vlm-f, erl-rl} require massive trajectory collection and repeated VLM queries to train a reward model aligned with LLM preferences. In contrast, VARL only queries VLMs at selected interaction steps, resulting in significantly lower computational cost.

\section{Related Work}

\subsection{VLM as Action Provider}

Recent advances have demonstrated that vision-language models (VLMs) can be leveraged as action providers, with most methods relying heavily on imitation learning. Early work in vision-and-language navigation (VLN) has shown that agents can be trained to follow natural language instructions in simulated environments \cite{vln_a_new_path,vln_do_v_improve_nv_agents,vln_adapt,vln_layout_aware_dreamer}. However, these methods generally assume the existence of teleportation-based actions, which limit their deployability in real-world embodied settings. Parallel research explores learning latent action vectors that serve as transferable representations across tasks and embodiments \cite{interleav_vla,uniVLA,vla_lapa}. Current progress in this direction has primarily focused on robotic manipulation, while its effectiveness in broader embodied environments remains less clear. Another branch of work focuses on training VLMs with behavior cloning from large-scale expert demonstrations \cite{vla_diffusion_vla,bc_z,interactive_language_real_time_talk,grounding_language_affordances_over_unstructured_data}. While such methods have proven effective, they typically require massive expert datasets and often rely on hierarchical strategies to decouple high-level planning from low-level motor control in order to improve success rates. More recent efforts attempt to utilize VLMs directly as autonomous agents capable of decision-making \cite{vlm_agent_adam,vlm_agent_li2024optimus,vlm_agent_fan2022minedojo,pi0}.
These VLM- and LLM-based agents can plan over high-level goals but still struggle to perform fine-grained low-level actions.
% In summary, most prior work in VLM as action providers has been rooted in imitation learning and behavior cloning, which inherently restricts performance to the quality and scale of the training dataset, making it difficult for such models to surpass expert demonstrations.

\subsection{VLM as Reward Designer} 
Another promising direction is to employ VLMs and LLMs as reward designers \cite{erl-rl,rl-vlm-f,ma2023eureka,t2r,language-2-reward}. Rather than providing direct actions, these models are used to shape or generate reward signals that guide reinforcement learning agents.
Specifically, Text2Reward (T2R) \cite{t2r} and Eureka \cite{ma2023eureka} generate reward functions by letting LLMs read task descriptions, and then mitigate errors caused by hallucinated reward signals through human interaction or genetic algorithms. In contrast, RL-VLM-F \cite{rl-vlm-f} and ERL-VLM \cite{erl-rl} adopt rating-based methods, where VLMs provide preference information over trajectories, which is then used to train more reliable reward functions.

\section{Preliminary}

\paragraph{Markov Decision Process} 
 
We represent each task as a Markov Decision Process (MDP) in the standard form: \(\mdp := \langle \sspace, \aspace, \rew, \trns, \gamma, \rho \rangle\), where \(\sspace\) and \(\aspace\) denote the state and action spaces, respectively, and we use \(\saspace := \sspace \times \aspace\) as shorthand for the joint state-action space. The reward function \(\rew \colon \saspace \to \dist([0,1])\) maps state-action pairs to distributions over the unit interval, while the transition function \(\trns \colon \saspace \to \dist(\sspace)\) maps state-action pairs to distributions over subsequent states. Lastly, \(\rho \in \dist(\sspace)\) represents the distribution over initial states. A (stochastic) policy \(\pi:\mathcal{S} \to \Delta(\mathcal{A})\) assigns a probability distribution over actions for each state, while a deterministic policy \(\mu:\mathcal{S} \to \mathcal{A}\) is the special case that selects a single action with probability 1.
 
\paragraph{Dataset} 
Let the standard replay buffer $\mathcal{D}$ store collected transitions of the form 
$(s, a, r, s', d)$, where $d \in \{0,1\}$ indicates whether the episode has ended. 
In addition, we maintain a heuristic buffer $\mathcal{D}_{\mathrm{llm}}$ that contains 
state–action pairs $(s, a_{\mathrm{llm}})$ generated by a vision-language model (VLM) generator 
$\mathcal{F} : (s,a) \mapsto a_{\mathrm{vlm}}$, which augments the transition with 
additional candidate actions. The overall training data distribution is thus derived 
from both $\mathcal{D}$ and $\mathcal{D}_{\mathrm{llm}}$.

To regulate when heuristic guidance is introduced, we define a \emph{trigger step set} 
$\mathcal{S}_h$, which specifies the training steps at which the VLM function 
$\mathcal{F}$ is queried to provide guidance. A weighting coefficient $\lambda$ is applied 
to balance the influence of heuristic actions on policy learning. Furthermore, a cutoff 
parameter $N_s$ is introduced, which determines the step after which heuristic actions 
are completely removed from the training process.

\section{Methods}

\begin{figure*}[htbp]
    \centering
    \includegraphics[width=1\textwidth]{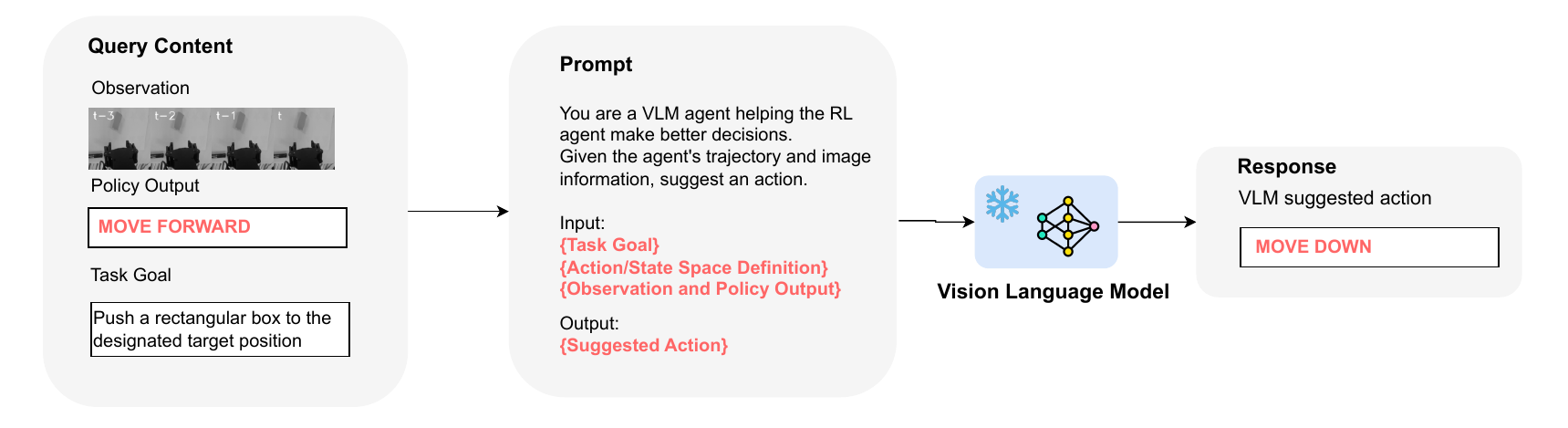}
    \caption{\textbf{VLM-based action generation process.} 
    The querying process of the VLM consists of two stages: 
    (1) collecting the transition from the agent’s trajectory, and 
    (2) using the transition information together with the prompt to query the VLM for a suggested action. 
    This enables the reinforcement learning agent to refine its policy by incorporating VLM-suggested actions.}
    % \label{fig:vlm_prompt}
    \label{fig:vlm_prompt}
\end{figure*}

\begin{algorithm}[t]
\caption{VARL}
\label{alg:policy_only_vlm_recent_rb}
\begin{algorithmic}[1]
\Require RL algorithm $\mathcal{A}$ with actor $\pi_\theta$ and critics $Q_{\phi_1}, Q_{\phi_2}$; replay buffer $\mathcal{D}$; guidance set $\mathcal{D}_{\text{llm}}$; trigger step set $S_h \subset \mathbb{N}$; recent-sample size $K$; entropy coefficient $\alpha$; \textbf{VLM-based action generator } $\mathcal{F}_{\text{vlm}}:(s,a)\mapsto a_{\text{vlm}}$
\State Initialize agent with $\mathcal{A}$; \quad $\mathcal{D}_{\text{llm}} \gets \emptyset$
\vspace{0.35em}

\Function{GetRecentSamples}{$\mathcal{D}, K$}
  \State \Return the $K$ most recent transitions $\{(s_i,a_i,r_i,s'_i,d_i)\}_{i=1}^K$ from $\mathcal{D}$
\EndFunction
\vspace{0.35em}

\For{global step $t = 1,2,\dots$}
  \State \textbf{Sample data:} $\mathcal{D} \gets (s_t,a_t,r_t,s_{t+1},d_t)$
  
  \If{$t \in S_h$}
     \State $\mathcal{B}_{\text{recent}} \gets \textsc{GetRecentSamples}(\mathcal{D}, K)$
     \For{each $(s,a,r,s',d) \in \mathcal{B}_{\text{recent}}$}
        \State $a_{\text{vlm}} \gets \mathcal{F}_{\text{vlm}}(s,a)$
        \State $\mathcal{D}_{\text{llm}} \gets \mathcal{D}_{\text{llm}} \cup \{(s, a_{\text{vlm}})\}$
     \EndFor
  \EndIf

  \State Update critics with standard soft TD update \cite{sac}
  \State Update policy with eq.(\ref{eq:actor_loss}) using $\mathcal{D}_{\text{llm}}$
\EndFor
% \vspace{-5pt}
\end{algorithmic}
\end{algorithm}

\subsection{VARL Framework}

The VARL framework consists of two key components: (i) a VLM-based action generator that produces heuristic actions from the buffer $\mathcal{D}$, and (ii) a policy shaping term that incorporates these actions into policy training, thereby biasing the agent’s exploration process.

As illustrated in Algorithm~\ref{alg:policy_only_vlm_recent_rb}, the VLM-based action generator periodically samples a batch of recent transitions, analyzes the states in conjunction with the task description, and generates heuristic actions, after which the resulting state–action pairs are incorporated into the policy training process.

\subsection{VLM-based Action Generator}

Formally, the VLM-based Action Generator is defined as a function $\mathcal{F}_{\text{vlm}}:(s,a)\mapsto a_{\text{vlm}}$, which takes a state-action pair as input and outputs a refined or alternative action suggestion. Unlike directly learning from all collected transitions, the generator is only invoked periodically during training: at specific intervals, a batch of recent transitions is sampled and passed to the VLM for guidance. This strategy significantly reduces the computational overhead of VLM inference, while still providing timely heuristic signals to the agent.

The generated state–action pairs $(s, a_{\text{vlm}})$ are stored in the guidance buffer $\mathcal{D}_{\text{llm}}$ and used to augment policy training. These heuristic actions expand the exploration space, increasing the chance of discovering novel trajectories that the policy alone might overlook. Compared with preference-based methods~\cite{rl-vlm-f,erl-rl}, which are restricted to choosing between pairs of trajectories, this approach provides broader heuristic signals and thus enhances sample diversity.

As shown in Figure~\ref{fig:vlm_prompt}, the effectiveness of the VLM-based Action Generator largely depends on the design of its prompts. Each prompt encodes task instructions, contextual information, and action-space constraints, following certain principles and a structured template to guide the generation of new actions. Well-crafted prompts ensure that the heuristic actions proposed by the VLM are both meaningful and consistent with the environment’s requirements.

\begin{figure*}[htbp]
    \centering
    \includegraphics[width=1\textwidth]{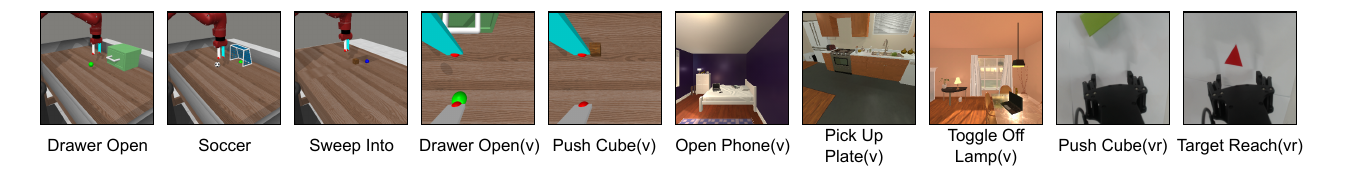}
        % \vspace{-5pt}
    \caption{\textbf{Evaluation environments.} VARL is evaluated in two simulators—\emph{Meta-World} (manipulation) and \emph{AI2-THOR} (navigation)—and in a real-world manipulation setup using a Realman Robotics RM-65B arm. The figure shows, from left to right, five Meta-World tasks, three AI2-THOR navigation tasks, and two real-world tasks. Abbreviations: (v) denotes vision-based tasks in simulation; (vr) denotes real-world vision-based tasks.}

    \label{fig:varl_tasks}
\end{figure*}

\subsection{Policy Shaping}
The policy shaping module augments standard policy iteration with heuristic state--action pairs $(s, a_{\text{llm}})$ from the VLM-based Action Generator. These pairs guide exploration in early training but are removed at a fixed step $N_s$, ensuring that the policy converge towards local optimal policy. This design both diversifies collected transitions and prevents the agent from being dominated by heuristic signals.

Formally, let $\pi_\theta(a \mid s)$ denote the policy parameterized by $\theta$, and let $Q_{\phi_1}, Q_{\phi_2}$ denote the critics. We define
\begin{equation}
Q(s,a) = \min(Q_{\phi_1}(s,a), Q_{\phi_2}(s,a)),
\end{equation}
the baseline policy loss
\begin{equation}
\mathcal{L}_{\pi} = 
\mathbb{E}_{s \sim \mathcal{D}} \Bigg[ 
\sum_{a \in \mathcal{A}} \pi_\theta(a \mid s) 
\Big( \alpha \log \pi_\theta(a \mid s) - Q(s,a) \Big) 
\Bigg],
\end{equation}
the behavior cloning loss
\begin{equation}
\mathcal{L}_{\text{BC}} = 
\mathbb{E}_{(s,a_{\text{llm}}) \sim \mathcal{D}_{\text{llm}}} 
\big[ - \log \pi_\theta(a_{\text{llm}} \mid s) \big],
\end{equation}
and the gating function
\begin{equation}
g(s,a_{\text{llm}}) \triangleq 
\mathbf{1}\!\left[a_{\text{llm}} \neq \arg\max_{a \in \mathcal{A}} Q(s,a)\right].
\end{equation}

With these definitions, the policy shaping term is written as
\begin{equation}
\mathcal{L}_{\text{ps}} 
= \lambda \, \mathbb{E}_{(s,a_{\text{llm}})\sim \mathcal{D}_{\text{llm}}}
\big[\, g(s,a_{\text{llm}})\, \big(-\log \pi_\theta(a_{\text{llm}}\mid s)\big) \big],
\end{equation}
where $\lambda$ controls the guidance weight. 
The gate disables the BC term precisely when $a_{\text{llm}}$ coincides with the critic-greedy action, preventing redundant reinforcement of the same action. Without this gating, the combined effect of $\mathcal{L}_{\pi}$ and $\mathcal{L}_{\text{BC}}$ would excessively concentrate probability mass on $a_{\text{llm}}$, leading to a rapid reduction in policy entropy and unstable critic learning.

For continuous action spaces, the baseline policy loss becomes:
\begin{equation}
\mathcal{L}_{\pi} = 
\mathbb{E}_{s \sim \mathcal{D}, a \sim \pi_\theta} 
\Big[ \alpha \log \pi_\theta(a \mid s) - Q(s,a) \Big].
\end{equation}
To gate the BC signal, let the policy distribution be Gaussian 
$\pi_\theta(\cdot \mid s)=\mathcal{N}(\mu_\theta(s), \Sigma_\theta(s))$. 
We define
\begin{equation}
\resizebox{0.9\linewidth}{!}{$
g(s,a_{\text{llm}}) \triangleq 
\mathbf{1}\!\left[\, 
(a_{\text{llm}}-\mu_\theta(s))^\top \Sigma_\theta(s)^{-1} (a_{\text{llm}}-\mu_\theta(s)) > \kappa^2
\,\right]
$}
\end{equation}
where $\kappa$ is a hyperparameter controlling the acceptance region. 
Thus, if the VLM action lies inside the high-probability ellipsoid 
$(a_{\text{llm}} \approx \mu_\theta(s))$, the BC term is disabled; 
otherwise, it is applied. The shaping loss $\mathcal{L}_{\text{ps}}$ is 
defined as in the discrete case but with this continuous gate.

The final actor loss combines the baseline policy update and the shaping term:
\begin{equation}
\label{eq:actor_loss}
\mathcal{L}_{\text{actor}} = \mathcal{L}_{\pi} + \mathbf{1}_{\{t \leq N_s\}} \, \mathcal{L}_{\text{ps}},
\end{equation}
where $\mathbf{1}_{\{t \leq N_s\}}$ activates the shaping term only during the first $N_s$ interactions to prevent overfitting to heuristic actions.

This policy shaping strategy leverages VLM guidance to improve early exploration while ensuring that the policy converges toward a locally optimal policy.

\subsection{Implementation Details}

We use Soft Actor-Critic (SAC)~\cite{sac} as the reinforcement learning (RL) solver for both discrete and continuous action settings, and adopt GPT-5~\cite{gpt5} as the vision-language model (VLM) to provide heuristic actions. The guidance weight is fixed at $\lambda = 10$ for all tasks, and the removal step is set to $N_s = 30{,}000$ for most tasks. Unless otherwise specified, the maximum training step is set to $500{,}000$.

\section{Experiments}

\begin{figure*}[htbp]
    \centering
    \includegraphics[width=\textwidth]{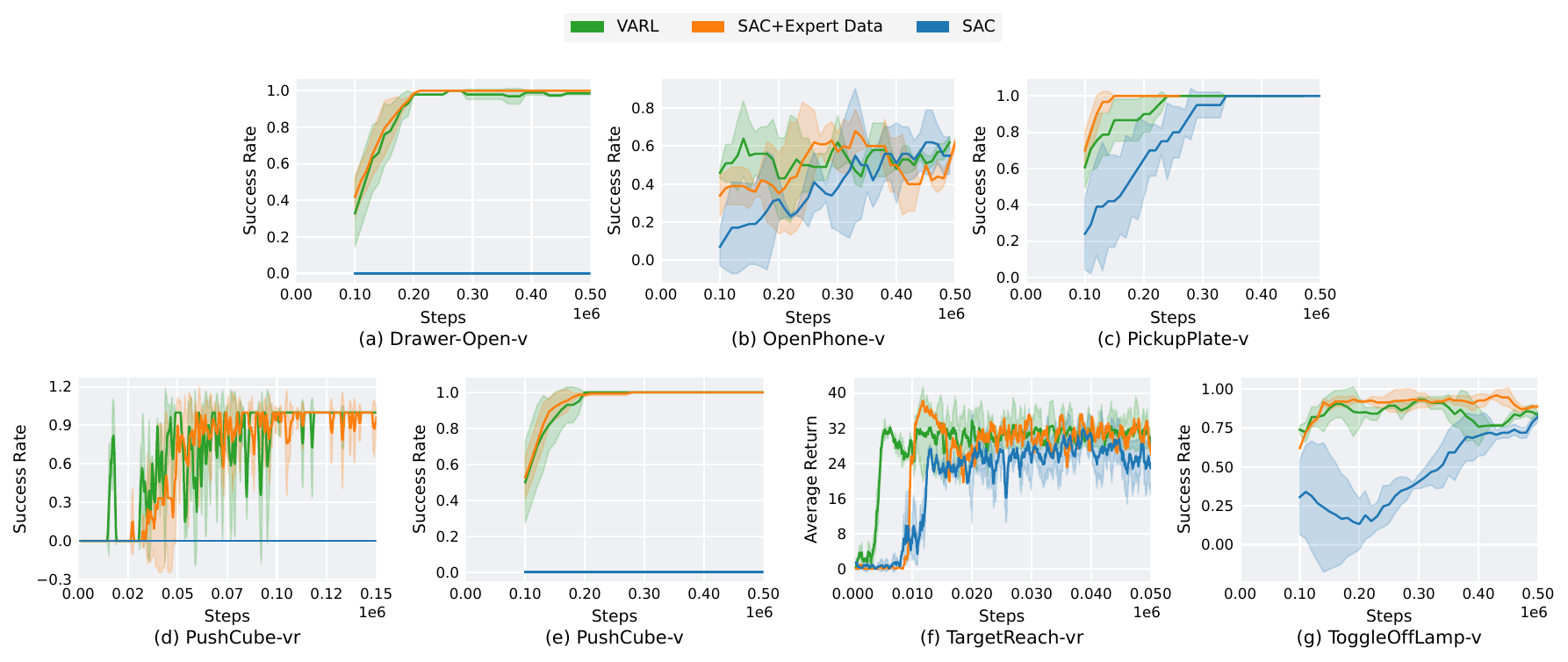}
        
    \caption{\textbf{Learning curves of different RL baselines across seven environments.} 
Each subplot corresponds to a distinct manipulation or navigation task: (a) Drawer-Open-v, (b) OpenPhone-v, (c) PickupPlate-v, (d) PushCube-vr, (e) Sweep-Target-v, (f) TargetReach-vr, and (g) ToggleOff-v. 
The x-axis denotes the number of environment interaction steps (in millions), while the y-axis shows either task success rate or average return depending on the task. 
Curves represent the mean performance of multiple runs, with shaded regions indicating one standard deviation. 
Comparison among \textit{VARL}, \textit{SAC+Expert Data}, and \textit{SAC} .}

    \vspace{-10pt}
    \label{fig:learning_efficiency}
\end{figure*}
 
We validate the effectiveness of VARL from two perspectives:
(1) \textbf{Generality:} assessing whether VARL can be applied across diverse task settings;
(2) \textbf{Improvement:} examining whether VARL can substantially enhance the performance of RL agents when provided with heuristic guidance from VLMs.

% \subsection{Environments and tasks}
\subsection{Environments and Tasks}
We evaluate VARL on ten tasks: three state-based manipulation tasks and two vision-based manipulation tasks from \emph{Meta-World}~\cite{mclean2025metaworldimprovedstandardizedrl}; three vision-based navigation tasks from \emph{AI2-THOR}~\cite{ai2thor}; and two real-world vision-based manipulation tasks. Tags: (v) = vision-based (simulation), (vr) = vision-based (real world).

\begin{enumerate}
    \item \textbf{Drawer Open} — Open a drawer; drawer pose randomized.  
In this task, the reward is changed to a designer-friendly reward: reward is given only when the drawer is open. 
    \item \textbf{Sweep Into} — Sweep a cube into a hole; cube pose randomized. 
    \item \textbf{Soccer} — Kick a ball into the goal; ball and goal poses randomized. 
    \item \textbf{Pick Up Plate (v)} — Navigate and pick up a plate once to complete the task. 
    \item \textbf{Open Phone (v)} — Navigate, pick up a phone, then open it on. 
    \item \textbf{Toggle Off Lamp (v)} — Navigate to a floor lamp and switch it off. 
    \item \textbf{Drawer Open (v)} — Open a drawer from image observations; drawer pose fixed.  
    In this task, the reward is changed to a designer-friendly reward: reward is given only when the drawer is open.  
    
    \item \textbf{Push Cube (v)} — Push a cube to a specified target location; cube pose fixed.  
    In this task, the reward is changed to a designer-friendly reward: reward is given only when the cube's position has changed. 
    \item \textbf{Target Reach (vr)} — Move to a specified target location in the real world.  
    Reward is given when the gripper closes on the target. 
    \item \textbf{Push Cube (vr)} — Push a cube to a specified target location in the real world.  
    Reward is determined only by the distance between the object and the target position.
\end{enumerate}

% \subsection{baselines}
\subsection{Baselines}

In the sample efficiency experiment, we compare VARL with SAC+expert data and vanilla SAC. SAC+expert data means that before training starts, SAC is provided with expert trajectories, which are added to the replay buffer to accelerate convergence and learning.  

In the comparison with reward shaping methods experiment, we compare VARL with RL-VLM-F~\cite{rl-vlm-f} and ERL-VLM~\cite{erl-rl}.

\subsection{Results and analysis}

\paragraph{VARL as a general framework for improving sample efficiency}  
As illustrated in Figure~\ref{fig:learning_efficiency}, VARL consistently improves sample efficiency across different task settings. By leveraging the biased action guidance provided by the VLM to guide exploration, VARL achieves significantly higher sample efficiency compared to vanilla SAC~\cite{sac}. When the environment provides only event-driven rewards, vanilla SAC struggles to learn because triggering such rewards through random exploration is extremely difficult. In contrast, VARL benefits from global observations obtained from the VLM, enabling it to explore in the directions suggested by the VLM. This makes VARL well suited for both sparse and event-driven reward settings, while also substantially reducing the need for manually designed reward functions.  

In dense-reward tasks, VARL can even outperform SAC initialized with expert trajectories. This advantage stems from the policy shaping mechanism, which enables rapid policy fitting. As a result, VARL converges faster in dense reward environments and can even surpass algorithms that rely on expert demonstrations.  

Overall, the combination of VLM-based action guidance and policy shaping prevents overfitting to biased actions while simultaneously exploiting VLM-provided exploration signals. This synergy makes VARL a general and effective online reinforcement learning framework that adapts seamlessly to diverse reward structures.

\begin{figure*}[t]
    \centering

    % 左边的图
    \begin{subfigure}[t]{0.38\textwidth}
        \centering
        \includegraphics[width=\textwidth]{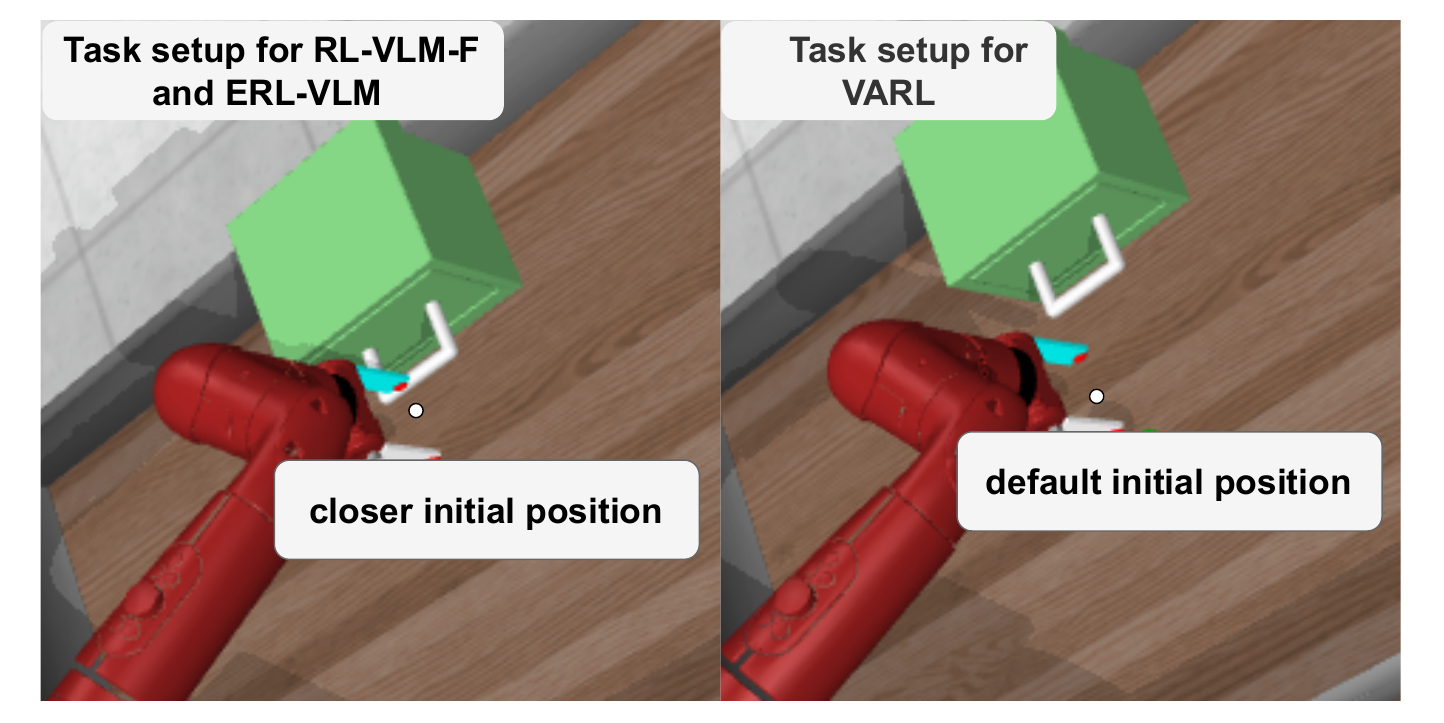}
        \caption{\textbf{VARL is evaluated under a more difficult task setup.} 
        We use the Drawer-Open environment as an example: \textsc{VARL} initializes from the default position, 
        while \textsc{RL-VLM-F} and \textsc{ERL-VLM} lower the difficulty by modifying the initial position 
        to be closer to the target.}
        \label{fig:task_setup}
    \end{subfigure}
    \hfill
    % 右边的图
    \begin{subfigure}[t]{0.6\textwidth}
        \centering
        \includegraphics[width=\textwidth]{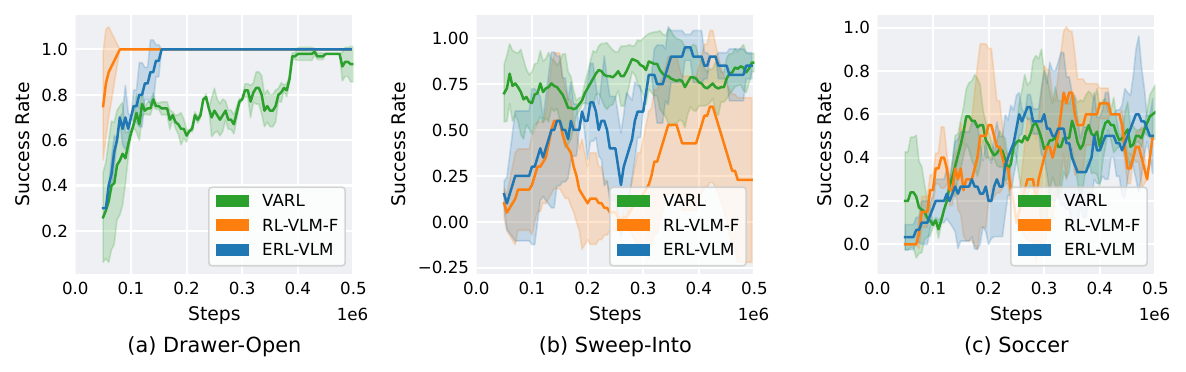}
        \caption{\textbf{Task success rates of \textsc{VARL}, \textsc{RL-VLM-F}, and \textsc{ERL-VLM} across three manipulation environments: (a) Drawer-Open, (b) Sweep-Into, and (c) Soccer.} Each curve shows the moving-average success rate (window size 10) aggregated over multiple training runs, with shaded regions denoting one standard deviation. The x-axis reports training steps (scaled by $10^6$), and the y-axis indicates the probability of successful task completion. }
        \label{fig:varl_vs_erlrl_rlvlmf}
    \end{subfigure}

\end{figure*}

% \begin{figure*}[t]
%     \centering

%     \begin{minipage}[t]{0.35\textwidth}
%         \centering
%         \includegraphics[width=\textwidth]{figures/task_setting_1.pdf}
%     \caption{\textbf{VARL is evaluated under a more difficult task setup.} 
%     We use the Drawer-Open environment as an example: \textsc{VARL} initializes from the default position, 
%     while \textsc{RL-VLM-F} and \textsc{ERL-VLM} lower the difficulty by modifying the initial position 
%     to be closer to the target.}
%         \label{fig:task_setup}
%     \end{minipage}%
%     % \hfill

%     \begin{minipage}[t]{0.6\textwidth}
%         \centering
%         \includegraphics[width=\textwidth]{figures/VLM_SAC_vs_RL_VLM_F_vs_ERL_RL_2025-09-16-05-55-59.pdf}
%         \caption{\textbf{Task success rates of \textsc{VARL}, \textsc{RL-VLM-F}, and \textsc{ERL-VLM} across three manipulation environments: (a) Drawer-Open, (b) Sweep-Into, and (c) Soccer.} Each curve shows the moving-average success rate (window size 10) aggregated over multiple training runs, with shaded regions denoting one standard deviation. The x-axis reports training steps (scaled by $10^6$), and the y-axis indicates the probability of successful task completion. }
%         \label{fig:varl_vs_erlrl_rlvlmf}
%     \end{minipage}
% \end{figure*}

\begin{table}[h]
\caption{\textbf{Comparison of VLM query usage.} 
All methods are trained for a maximum of 500k steps. 
We report the number of VLM queries, the minibatch size used per query, 
and the resulting total number of referenced samples (\#Queries~$\times$~mbsize). 
Results highlight the drastic reduction of VLM calls in \textsc{VARL}.}
\label{tab:vlm_query_usage}
\begin{center}
\begin{tabular}{|c||c|c|c|}
\hline
\textbf{Method} & \textbf{\#Queries} & \textbf{mbsize} & \textbf{Total samples} \\
\hline
ERL-VLM     & 5000 & 50  & 250{,}000 \\
\hline
RL-VLM-F   & 5000 & 128 & 640{,}000 \\
\hline
VARL       & 3    & 500 & 1{,}500 \\
\hline
\end{tabular}
\end{center}
\end{table}

\paragraph{VARL requires low computational resources and achieves superior performance compared to VLM-based reward shaping methods}  
VARL and VLM-based reward shaping approaches represent two distinct paradigms. While VARL aims to improve sample efficiency through policy shaping with VLM-guided action fusion, RL-VLM-F~\cite{rl-vlm-f} and ERL-VLM~\cite{erl-rl} instead modify the reward function to improve the success rate of the RL agent. To ensure a fair comparison, we adopt designer-friendly rewards for evaluation, which increases the learning difficulty, and initialize tasks from more challenging configurations (Figure~\ref{fig:task_setup}).  

As shown in Figure~\ref{fig:learning_efficiency}, VARL achieves comparable performance to reward shaping methods in two environments, and even surpasses them in the Sweep-Into task. This demonstrates that leveraging VLM information through policy shaping is more effective than reward shaping. Reward shaping methods require periodic updates to the reward model, forcing the agent to discard previously learned policies and restart training. This process leads to a substantial number of additional VLM queries, increased policy updates, and significant computational overhead. In contrast, VARL drastically reduces VLM query usage (Table~\ref{tab:vlm_query_usage}) while maintaining or surpassing performance.  

These results highlight the efficiency and generality of VARL: the combination of policy shaping and minimal VLM interaction provides a lightweight yet powerful framework that outperforms reward shaping methods both in computational cost and in task success rates.

\paragraph{VARL enables online reinforcement learning in the real world}  
VARL is a novel framework that integrates VLM heuristics through policy shaping to improve sample efficiency, and its effectiveness extends beyond simulation to real-world settings. In prior work, real-world reinforcement learning has predominantly relied on human-in-the-loop training~\cite{hil_serl}, behavior cloning or offline RL~\cite{chen2025conrft}, VLA fine-tuning~\cite{pi0,uniVLA}, or sim-to-real transfer~\cite{google_soccer,drone_racing}. While these approaches have achieved progress, they invariably depend on expert demonstrations or detailed environment modeling. In scenarios where demonstrations or accurate models are unavailable, such methods become impractical. A more general solution is to directly perform online reinforcement learning in novel environments. However, this paradigm has remained underexplored due to the challenges of low sample efficiency and the difficulty of designing effective reward functions.  

To address these limitations, we directly deploy VARL on a RM-65B robotic arm and conduct online reinforcement learning from scratch in two real-world tasks. For reward design, we adopt user-friendly signals; for example, in the Push-Cube task, the reward is solely based on the distance between the object and the target location.  

As shown in Figure~\ref{fig:learning_efficiency}, VARL successfully learns to reach specific target positions within approximately 3,000 interactions, and acquires the ability to push a cube to the designated location within 30,000–50,000 interactions. These results demonstrate that VARL makes online reinforcement learning feasible in real-world robotic scenarios, even under sparse or minimally designed reward signals.

\begin{figure}[htbp]
    \centering
    \includegraphics[width=\columnwidth]{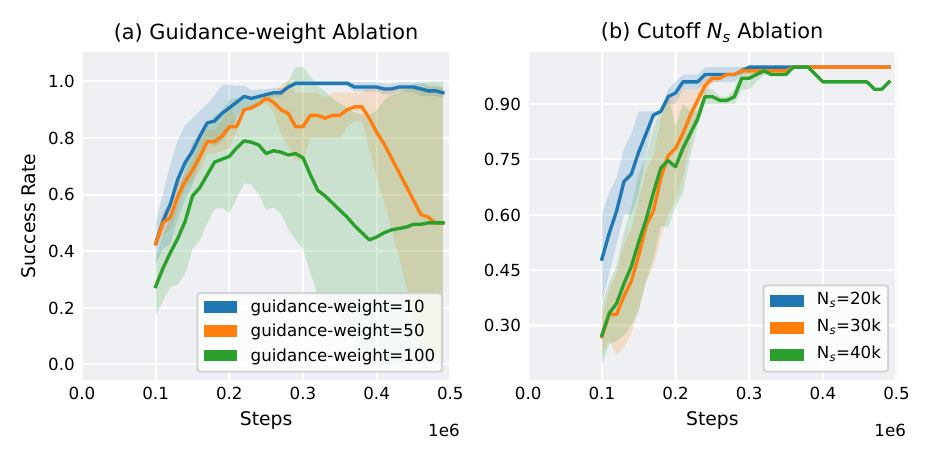}
        
    \caption{\textbf{Sensitivity analysis of hyperparameters.} Success rate curves are reported under different settings of \textit{guidance-weight} (left) and cutoff step $N_s$ (right). The results show that VARL is not highly sensitive to these hyperparameters, indicating that it does not require frequent tuning across tasks and thus exhibits stronger robustness.}
    \vspace{-10pt}
    \label{fig:ablation_study_varl}
\end{figure}

\subsection{Ablation Study}
In this section, we conduct sensitivity analyses to better understand the influence of key hyperparameters in VARL. Specifically, VARL introduces two important factors that determine the strength and duration of policy shaping:

\begin{enumerate}
    \item \textbf{Coefficient $\lambda$}: controls the magnitude of the heuristic action term.
    \item \textbf{Fixed removal step $N_s$}: determines the length of the shaping period.
\end{enumerate}

We evaluate agent performance under different values of $\lambda$ and $N_s$ on a representative vision-based task, \textit{drawer-open (v)}.

As illustrated in Figure~\ref{fig:ablation_study_varl}, the sensitivity analysis reveals several insights:

\textbf{Effect of $\lambda$ (Guidance-weight Ablation, Fig.~\ref{fig:ablation_study_varl}a):} Increasing $\lambda$ enhances the contribution of heuristic guidance. A moderate value (e.g., $\lambda=50$) yields the highest success rate, balancing external guidance with the agent’s own exploration. However, overly large values (e.g., $\lambda=100$) suppress autonomous learning and lead to noticeable performance degradation, even though the training process remains stable without collapse.

\textbf{Effect of $N_s$ (Cutoff Ablation, Fig.~\ref{fig:ablation_study_varl}b):} Varying $N_s$ controls how long heuristic shaping persists. Smaller values (e.g., $N_s=20k$) reduce early-stage benefits but allow faster convergence to an optimal long-term policy. Larger values (e.g., $N_s=40k$) overly bias the agent toward heuristic actions, which can hinder its ability to refine strategies beyond the provided guidance. Nevertheless, the differences across settings remain relatively small, indicating that VARL can adapt across a range of shaping durations.

Overall, the results demonstrate that VARL is not highly sensitive to either $\lambda$ or $N_s$, as performance curves remain robust across a broad range of values. This robustness suggests that the proposed method does not require precise hyperparameter tuning, making it more practical for deployment in diverse tasks.

\section{Conclusion}

In this work, we proposed \textbf{VARL}, a framework that improves the sample efficiency of online reinforcement learning (RL) agents by querying vision-language models (VLMs) for heuristic actions. 
The main innovations of VARL lie in two aspects:  
(1) algorithmically, we introduce a gating function to constrain the policy from directly overfitting to VLM-provided actions, thereby preventing divergence of the Q-function;  
(2) experimentally, we demonstrate that VARL enables direct online learning in physical environments without requiring expert demonstrations, highlighting its potential for self-improving agents.  

Compared to rating-based reward shaping methods, VARL provides richer action diversity while consuming fewer computational resources and without relying on the accuracy of VLM outputs. 
Moreover, VARL preserves the optimality of the underlying RL agent by avoiding reward bias.  

Our experiments demonstrate that VARL serves as a general framework applicable across diverse task settings, consistently enhancing sample efficiency. 
In addition, its lightweight design makes it suitable for direct deployment in physical environments for online RL.  

For future work, we plan to extend VARL to more complex real-world tasks and investigate the integration of video-generation techniques to further enrich heuristic guidance. 
We will also aim to scale its application to broader embodied environments, extending beyond robotic manipulation.

%%%%%%%%%%%%%%%%%%%%%%%%%%%%%%%%%%%%%%%%%%%%%%%%%%%%%%%%%%%%%%%%%%%%%%%%%%%%%%%%
\bibliography{references_this_paper}
\bibliographystyle{unsrt}

% \begin{thebibliography}{99}

% \bibitem{c1} G. O. Young, ÒSynthetic structure of industrial plastics (Book style with paper title and editor),Ó 	in Plastics, 2nd ed. vol. 3, J. Peters, Ed.  New York: McGraw-Hill, 1964, pp. 15Ð64.
% \end{thebibliography}

\end{document}